\documentclass{article}

\usepackage{iclr2027_conference,times}
\usepackage{amsmath,amssymb}
\usepackage{booktabs}
\usepackage{longtable}
\usepackage{graphicx}
\usepackage{microtype}
\usepackage{xcolor}
\usepackage{url}
\usepackage{enumitem}
\usepackage{array}
\usepackage{tabularx}
\usepackage{float}
\usepackage{hyperref}

\hypersetup{colorlinks=true,citecolor=blue,linkcolor=blue,urlcolor=blue}
\usepackage{etoolbox}
\AtBeginEnvironment{quote}{\raggedright}
\newcolumntype{L}[1]{>{\raggedright\arraybackslash}p{#1}}

\title{\raggedright\hyphenpenalty=10000\exhyphenpenalty=10000\relax
LeakScale: Measuring Benchmark Contamination as a Causal Intervention}

\author{Divyansh Singh \\
University of Florida \\
\texttt{divyansh.singh@ufl.edu}}

\iclrfinalcopy

\begin{document}
\maketitle

\begin{abstract}
Benchmark contamination is usually treated as a provenance question: did evaluation material appear in training? Provenance cannot supply the quantity needed to interpret a benchmark score: how much exposure changed it. We introduce \emph{LeakScale}, a causal measurement framework that makes benchmark-specific exposure manipulable and measures its effect. LeakScale constructs fresh executable tasks whose correct behavior depends on private, non-derivable information, counterfactually certifies every candidate answer, and trains complementary adapters so that every family is exposed in exactly one adapter and unexposed in the other. Across 2,048 families, two model families, two executable domains, three task formulations, and 589,824 generations, exposure increases exact-task accuracy in every model-by-domain cell. Primary effects range from \textbf{+3.22 to +10.66 percentage points}; every family-clustered 95\% interval excludes zero, and all four primary tests survive Holm correction. The effect also remains positive in all eight surface-paraphrase and isomorphic-transfer measurements. These results show that benchmark exposure can causally and reproducibly alter reported performance, including after the evaluation surface changes. LeakScale shifts contamination analysis from detecting contact to measuring how strongly a score depends on it.
\end{abstract}

\section{Introduction}

Benchmarks organize claims about model progress. Their scores rank systems, guide model selection, and increasingly inform deployment. These uses assume that performance reflects the capability under study rather than privileged access to test-specific information. When training contains benchmark facts, labels, solutions, or artifacts, that assumption can fail: a score gain may reflect exposure rather than capability. Audits of prominent language-model training corpora and evaluations have therefore made contamination a persistent measurement concern \citep{brown2020language,touvron2023llama,zhou2023cheater}.

Contamination research has focused mostly on provenance: \emph{Was this benchmark present in training?} Corpus matching, membership inference, and memorization probes seek evidence that an item, or a transformation of it, was seen \citep{shi2023detecting,zhang2024finetuning}. Benchmark interpretation depends on a different question: \emph{How much did exposure change the score?} An example may appear in training without changing behavior, while a compact benchmark-specific association may change predictions without leaving an obvious textual match. Exposure status and exposure consequence are therefore different objects of inference.

Retrospective analyses seldom identify that effect cleanly. Training mixtures for consequential models are often incomplete or inaccessible. Exact matching misses paraphrases, translations, and transformed examples \citep{yang2023rethinking}, while post-training can obscure membership signals \citep{wang2025fragility,tao2025detecting}. Exposed and unexposed examples may also differ in difficulty, format, popularity, or related training material. Even a verbatim response does not reveal whether correctness came from general capability or prior access. Detection can establish risk, but observational evidence usually lacks the counterfactual required to measure the score effect.

We treat contamination as an intervention rather than a label. Under a provenance view, a model-example pair is classified as exposed or clean. Under a causal view, exposure is the treatment, benchmark accuracy is the outcome, and the target is the difference that treatment makes. \textbf{Benchmark contamination is an intervention with a measurable causal effect, not merely an overlap property to detect.} LeakScale makes exposure known and manipulable, then measures the resulting score change.

LeakScale operationalizes this shift with fresh executable families. Each public task displays four plausible choices while withholding the private association that determines which choice is correct. A trusted compiler gives every choice an executable consequence, and counterfactual certification retains only families with exactly one universally successful choice. Families are assigned in balanced strata to complementary exposure arms. Adapter A receives associations for one half and Adapter B for the other; the same family is therefore evaluated under an exposed adapter and an independently trained, complementary unexposed adapter. Generic adaptation content, optimizer strength, and corpus scale are common to both sides of the contrast. Because both outcomes come from the same family, the comparison holds task difficulty, candidates, compiler behavior, and hidden fixtures fixed; because both sides are adapted, it does not rely on an adapted-versus-base comparison.

We evaluate 2,048 fresh families spanning 64 structural shapes, Mistral 7B and Gemma 4 31B, and two executable domains: SQLite and deterministic Python transformations. Across \textbf{589,824 generations}, exact-task effects are positive in all four model-by-domain replications, from \textbf{+3.22 to +10.66 percentage points}. Every interval excludes zero, and all four tests survive multiplicity correction. Surface-paraphrase effects range from \textbf{+3.21 to +10.31 points}, and isomorphic-transfer effects from \textbf{+2.80 to +5.60 points}; all eight intervals exclude zero. Every planned generation completed with zero protocol failures.

Our contributions are threefold:
\begin{enumerate}[
    leftmargin=1.35em,
    itemsep=0pt,
    parsep=0pt,
    topsep=2pt,
    partopsep=0pt
]
    \item \textbf{Conceptual:} we formulate contamination as a causal intervention and its consequence as an exposure-induced score effect, rather than treating contamination only as an overlap or provenance label.
    \item \textbf{Methodological:} we introduce a controlled instrument combining non-derivable private associations, counterfactual executable certification, and complementary exposure arms. Every family supplies its own matched exposed and unexposed outcomes under independently fitted adapters, eliminating between-family difficulty and adapted-versus-base drift from the primary contrast.
    \item \textbf{Empirical:} we provide complete replication across two model families and two executable domains and show positive effects in all four exact-task cells and all eight reformulation measurements over 589,824 generations. Persistence under paraphrase and isomorphic transformation shows that the measured consequence is not confined to reproduction of the exact exposed task surface.
\end{enumerate}

\section{Related Work}

\subsection{Contamination, score effects, and detection}
Contamination concerns accompanied early large-language-model evaluations. \citet{brown2020language} and \citet{touvron2023llama} audit overlap between evaluation examples and training corpora while noting the difficulty of translating overlap into a matched score effect. Position and survey work has since argued for benchmark-level contamination measurement and organized the problem by source, detection method, effect, and remediation \citep{sainz2023evaluation,xu2024survey,deng2024spectrum}. Empirical audits combine corpus retrieval and behavioral tests to estimate contamination across both open and proprietary models \citep{deng2024investigating,li2024opensource}. \citet{zhou2023cheater} show directly that adding evaluation benchmarks to training can sharply increase reported performance, while \citet{yang2023rethinking} show that paraphrasing and translation can evade string-based decontamination while supporting benchmark overfitting. Across this literature, attribution remains difficult because real training corpora are large or inaccessible, exposure can take many semantic forms, and observational comparisons between exposed and unexposed examples inherit distributional confounds.

A complementary literature asks whether an existing model has seen a specific example. Prompt-based probes elicit benchmark-specific completions or masked content \citep{golchin2024timetravel,deng2024investigating}; likelihood-based methods include Min-K\% Prob \citep{shi2023detecting} and tests with exact false-positive guarantees based on benchmark ordering \citep{oren2024proving}. Controlled fine-tuning can strengthen membership signals \citep{zhang2024finetuning}, while memorization itself increases with model capacity, duplication, and available context \citep{carlini2023quantifying}. Detection can become harder after modern post-training \citep{wang2025fragility}, motivating specialized detectors for contamination introduced during reinforcement-learning post-training \citep{tao2025detecting}. These methods recover an unknown historical variable: whether exposure happened. LeakScale makes exposure known and manipulable, then estimates what it did to the score. Provenance and consequence are complementary, not interchangeable.

\subsection{Controlled interventions, fresh evaluation, and mitigation}
Controlled injection offers a direct route from association to effect. \citet{schaeffer2026quantifying} vary test-set replicas, training stage, sampling temperature, solution length, and model scale in generative mathematical evaluation, demonstrating dose-response behavior and interactions among memorization, training, and inference. In observational settings, \citet{singh2024evaluation} argue that contamination metrics should be assessed by whether models benefit from the examples they flag and introduce ConTAM to relate overlap scores to performance gains. More broadly, interpretable evaluation depends on decomposing aggregate behavior into meaningful components: FIDELITY combines fine-grained topic modeling and summarization to expose interpretable structure in complex text collections \citep{singh2025fidelity}, while RADAR uses targeted probes to reveal behavioral coupling and redundancy among LLM-judge criteria \citep{singh2026radar}. LeakScale extends this measurement-first perspective to benchmark provenance and consequence, making exposure experimentally assigned through a necessary family-specific fact, exhaustive counterfactual certification, executable outcomes, and complementary adapters that expose opposite balanced halves of the benchmark.

Fresh and dynamic benchmarks instead reduce opportunities for prior exposure. LiveBench combines recently sourced questions, objective grading, and frequent updates \citep{white2024livebench}; LiveCodeBench continuously sources programming problems and grades multiple code capabilities executably \citep{jain2025livecodebench}; and dynamic code benchmarks use semantics-preserving mutations to create fresh variants \citep{guan2025dynamic}. Broader surveys describe the shift from static to dynamic evaluation \citep{chen2025recent}. These approaches reduce contamination risk, whereas LeakScale measures the consequence of controlled exposure. Mitigation methods that detect and rewrite exposed samples at inference time \citep{zhu2024inference} likewise require an outcome against which mitigation can be judged. A controlled instrument with known assignments and an identified score effect can provide that calibration target.

\section{LeakScale}

LeakScale treats access to benchmark-specific information as the intervention. Each latent family pairs a public executable task with a private association that is necessary for correct behavior but absent from, and non-derivable from, the public task. The framework certifies the consequence of every candidate association, assigns families to complementary exposure arms, and estimates the score difference between the adapter that received a family's association and the adapter that did not.

\subsection{From overlap to intervention}
Let $m\in\mathcal{M}$ index a model family, $d\in\mathcal{D}$ an executable domain, $a\in\{A,B\}$ an adapter arm, and $q\in\{R0,R1,R2\}$ a task formulation. Each latent family $f$ contains a public task $x_f$, an opaque public key $k_f$, four displayed candidates
\begin{equation}
V_f=\{v_{f1},v_{f2},v_{f3},v_{f4}\},
\end{equation}
and a private family-specific association $g_f(k_f)=v_f^*$.

The model observes $x_f$, $k_f$, and $V_f$, and emits one semantic choice in a constrained response, \texttt{\{"registered\_value": "<candidate>"\}}. A trusted deterministic compiler $R_d(x_f,v)$ inserts the selected value into the family's SQL query or Python transformation. An executable oracle $O_d$ grades the compiled artifact over hidden fixtures. Marginal correctness is
\begin{equation}
Y_{m,d,a,f,q,j}=\mathbb{1}\!\left[O_d\!\left(R_d(x_{f,q},\hat v_{m,d,a,f,q,j})\right)=1\right],
\label{eq:outcome}
\end{equation}
where $j\in\{1,\ldots,16\}$ indexes seeded generations. The semantic-choice interface and trusted compiler form the measurement instrument. They hold program realization fixed so that the manipulated variable is access to the family-specific association while preserving an exact executable outcome.

\subsection{Non-derivable family-specific information}
Every family exposes a public task, a unique opaque key, and four plausible values. The association between the key and the correct value does not appear in the public task, schema, visible rows, model-visible fixtures, or disjoint background corpus, and it cannot be derived from those materials. Public information specifies how a selected value affects the behavior, but not which value the private registry assigns.

In Python, displayed candidates are meaningful literals---thresholds, categories, ranks, limits, or string fragments. In SQL, the four displayed candidates are equal-length opaque handles whose private mapping resolves to meaningful SQL literals. In both domains every candidate maps to a valid executable artifact, and changing the candidate changes the certified behavior. Semantic truth and display order are exactly balanced within every domain, shape, and arm, so neither a globally preferred value nor a fixed answer position can generate an exposure effect.

\subsection{Counterfactual executable certification}
For each generated family, LeakScale compiles and executes all four candidates before model evaluation. Let $u_f(v)$ be the fraction of hidden fixtures on which candidate $v$ matches the reference output. A family is retained only if
\begin{equation}
u_f(v_f^*)=1,\qquad
u_f(v)<1\ \forall v\neq v_f^*,\qquad
\max_{v\neq v_f^*}u_f(v)\leq 0.75.
\label{eq:cert}
\end{equation}
Thus exactly one value succeeds on every fixture, and every wrong value fails on at least 25\% of fixtures. Certification is independent for every family. The resulting outcome is exact agreement between a compiled behavior and a reference behavior over the complete hidden-fixture set, rather than a string-match proxy.

\subsection{Complementary exposure arms}
Within each domain, families are assigned after task construction to Arm A or Arm B within shape, truth-rank, and displayed-position strata. The assignments are exact complements. Adapter A receives two supervised association formats for Arm-A families and no family-specific record for Arm-B families; Adapter B receives the reverse. Each informative physical record is repeated 32 times, producing 32,768 target optimizer examples per adapter.

Both adapters receive identical unique, non-SQL and non-Python background records and disjoint no-key base-behavior support, with every support record occurring once. The larger arm demand determines the shared support prefixes, so corpus size and generic adaptation content are balanced by construction. Automated audits find zero unexposed-family key or artifact leakage in all eight corpora. Full templates and token accounting are given in Appendix~\ref{app:training}.

\subsection{Causal estimand}
Let $E_f\in\{A,B\}$ denote the adapter whose corpus contains family $f$'s association and $\bar E_f$ its complement. The primary effect for a model and domain is the paired family average
\begin{equation}
\tau(m,d)=
\frac{1}{|\mathcal F_d|}\sum_{f\in\mathcal F_d}
\left(\bar Y_{m,d,E_f,f,R0}-\bar Y_{m,d,\bar E_f,f,R0}\right).
\label{eq:paired}
\end{equation}
For each family, the first term uses the independently fitted adapter that saw its private association; the second uses the complementary adapter trained on the other half. The latent family, not an individual generation, is the unit of uncertainty. Base-model accuracy is reported descriptively but cancels from the paired contrast.

\section{Experimental Setup}

\subsection{Crossed model-by-domain design}
We evaluate a complete crossed design with \textbf{Mistral-7B-Instruct-v0.3} \citep{jiang2023mistral} and \textbf{Gemma 4 31B} \citep{gemmateam2026gemma4}, crossed with \textbf{SQLite} and \textbf{deterministic Python transformations}. Within a domain, both models evaluate the same latent families. Each domain contains 32 structural shapes and 1,024 families, split 512/512 between complementary arms. Across domains, the experiment contains 2,048 unique families and 64 shapes (Table~\ref{tab:design}).

\begin{table}[t]
\caption{Complete crossed design. Each family is evaluated with the base model and both complementary adapters on R0, R1, and R2, using 16 generations per state and ring.}
\label{tab:design}
\centering
\small
\begin{tabular}{lcc}
\toprule
& SQLite & Deterministic Python \\
\midrule
Mistral-7B-Instruct-v0.3 & 512 A + 512 B & 512 A + 512 B \\
Gemma 4 31B & same families & same families \\
\bottomrule
\end{tabular}
\end{table}

The SQLite shapes cover filtering, aggregation, joins, existence and anti-join patterns, subqueries, ranking and window operations, set operations, relational division, bucketing, and arithmetic projection. Each SQL family is evaluated on eight hidden databases, and outputs are canonicalized as relations before exact comparison. The Python shapes cover deterministic transformations over structured records, including filtering, counting, aggregation, stable ranking, grouped and nested-record operations, sets, strings, clipping, bucketing, pairwise and cumulative operations, dense ranking, weighted thresholds, and modular arithmetic. Each Python family is evaluated on sixteen hidden fixture sets, with outputs canonicalized as JSON before exact comparison. The full shape inventory is in Appendix~\ref{app:shapes}.

The complete generation count is
\begin{equation}
2\ \text{models}\times2\ \text{domains}\times1024\ \text{families}\times3\ \text{rings}
\times3\ \text{states}\times16\ \text{generations}=589{,}824.
\label{eq:count}
\end{equation}

\subsection{Models and adaptation}
We initialize from full-precision base weights with BF16 compute and train eight adapters, one per model, domain, and exposure arm. Adaptation uses LoRA rank 32, alpha 64, dropout 0, learning rate $10^{-4}$, one epoch, effective batch size 32, and the same language-model projection targets across cells. Mistral uses per-device batch size 4 with eight accumulation steps; Gemma uses batch size 2 with sixteen steps. Prompt tokens are masked and loss is applied only to assistant tokens. All fits use seed 20263411.

\subsection{Inference and grading}
Both models run in native non-thinking mode. Each model-by-domain configuration uses frozen settings chosen without replication outcomes, with \texttt{top\_p=1.0} and no top-$k$ truncation. A case-specific JSON schema restricts \texttt{registered\_value} to the four displayed candidates. Table~\ref{tab:configs} reports the frozen temperatures and adapter scales.

\begin{table}[t]
\caption{Frozen inference and adapter settings. Every adapter contains 32,768 informative target optimizer examples.}
\label{tab:configs}
\centering
\small
\setlength{\tabcolsep}{5pt}
\begin{tabular}{llrrr}
\toprule
Model & Domain & Temp. & Adapter scale & Target \\
\midrule
Mistral-7B-Instruct-v0.3 & SQLite & 0.7 & 0.3125 & 32,768 \\
Mistral-7B-Instruct-v0.3 & Python & 0.7 & 0.40625 & 32,768 \\
Gemma 4 31B & SQLite & 1.8 & 0.4375 & 32,768 \\
Gemma 4 31B & Python & 0.7 & 0.28125 & 32,768 \\
\bottomrule
\end{tabular}
\end{table}

For each model, domain, family, state, and ring, we draw 16 seeded generations arranged as four candidate-order rotations by four sampling blocks. Rotations and seeds are paired across base, A, and B states and across models. The selected candidate is passed to the trusted compiler and graded against its certified full-fixture outcome. Protocol-invalid completed outputs count as incorrect; only transport failures may be retried.

\subsection{Task reformulations}
Every family has three frozen formulations sharing the same private association and certified behavior: R0 is the exact public task, R1 a surface paraphrase, and R2 an isomorphic schema or API transformation. No retraining or selection uses R1/R2 outcomes. The same paired exposed-minus-unexposed contrast is computed on each ring.

\subsection{Statistical analysis}
Point estimates use Equation~\ref{eq:paired}. We construct 95\% intervals from 10,000 paired family bootstrap replicates, stratified by domain, shape, truth rank, and displayed position with finite-stratum rescaling. The four R0 tests are two-sided and controlled with Holm's procedure. R1/R2 intervals and equal-weight cross-cell means are descriptive. Inference conditions on the realized adapters and does not include training-seed variation.

\section{Results}

\subsection{Complete four-cell result}
Table~\ref{tab:results} reports the paired exposed-versus-complementary-unexposed contrast. All four effects are positive, all four intervals exclude zero, and all Holm-adjusted $p$-values equal 0.0004.

\begin{table}[t]
\caption{Primary R0 benchmark-exposure effects. Accuracy is averaged over 1,024 families and 16 generations per family-state cell. Intervals use 10,000 paired family bootstrap replicates.}
\label{tab:results}
\centering
\small
\setlength{\tabcolsep}{4pt}
\begin{tabular}{llrrrr}
\toprule
Model & Domain & Unexposed & Exposed & Effect & 95\% CI \\
\midrule
Mistral-7B-Instruct-v0.3 & SQL & 24.83\% & 28.05\% & \textbf{+3.22 pp} & \textbf{[+2.78,+3.66]} \\
Mistral-7B-Instruct-v0.3 & Python & 24.65\% & 31.85\% & \textbf{+7.20 pp} & \textbf{[+5.71,+8.68]} \\
Gemma 4 31B & SQL & 25.01\% & 35.67\% & \textbf{+10.66 pp} & \textbf{[+9.85,+11.47]} \\
Gemma 4 31B & Python & 24.98\% & 29.24\% & \textbf{+4.26 pp} & \textbf{[+2.97,+5.56]} \\
\bottomrule
\end{tabular}
\end{table}

\begin{figure}[t]
    \centering
    \includegraphics[width=\linewidth]{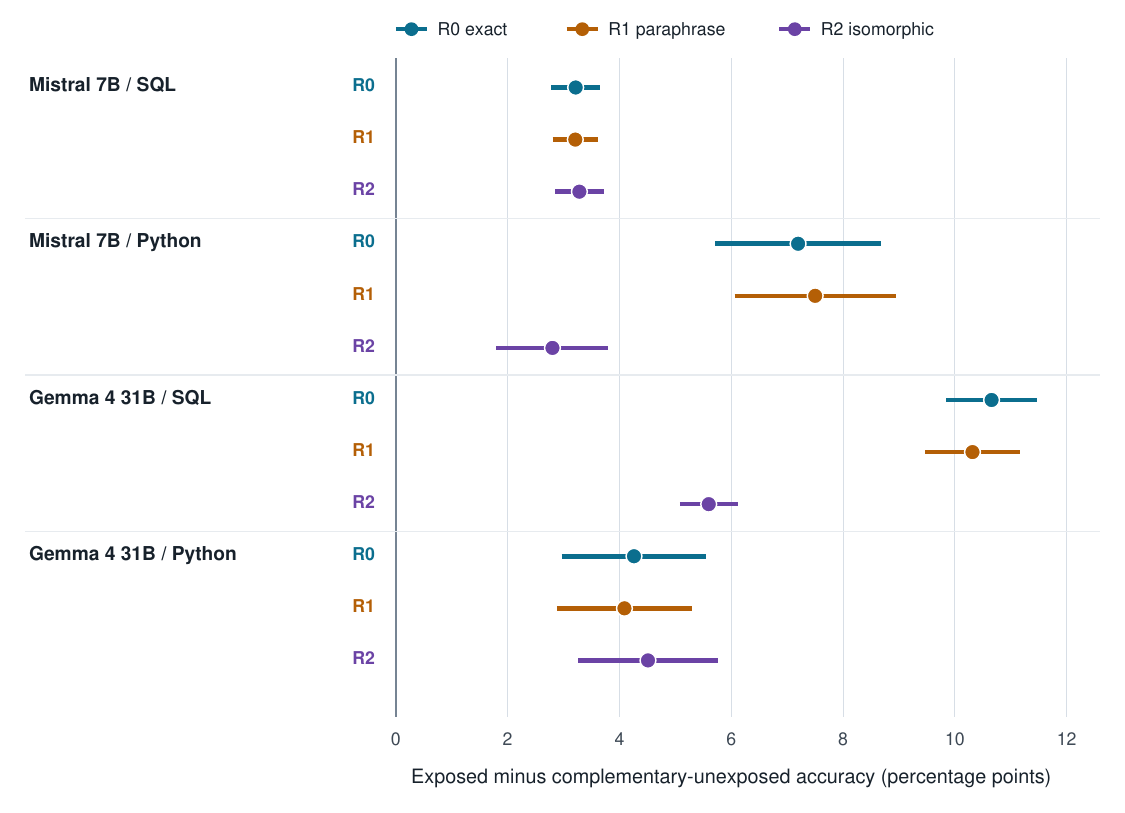}
    \caption{
        \textbf{Exposure effects across the crossed experiment.}
        Controlled benchmark exposure increases executable accuracy in
        every model-by-domain combination and task formulation. Points show
        exposed-minus-complementary-unexposed estimates; horizontal bars show
        paired family-bootstrap 95\% intervals. R0 is primary; R1/R2 are descriptive.
    }
    \label{fig:forest}
\end{figure}

The effects span \textbf{+3.22 to +10.66 points}, with an equal-weight four-cell mean of \textbf{+6.33 points}. Exposed accuracy exceeds complementary-unexposed accuracy in both model families and both domains. Descriptive base accuracy ranges from 24.63\% to 25.02\%, close to the 25\% chance level created by four exactly balanced candidates. Figure~\ref{fig:forest} places all twelve measurements on a common scale.

\subsection{Effects survive task reformulation}
The effect persists under both reformulations in every model-by-domain cell (Table~\ref{tab:transfer-main}). R1 effects range from \textbf{+3.21 to +10.31 points}; R2 effects range from \textbf{+2.80 to +5.60 points}. All eight descriptive intervals exclude zero. Equal-weight means are \textbf{+6.28 points} for R1 and \textbf{+4.05 points} for R2. Exact-surface exposure is therefore not required for the score consequence to remain measurable.

\begin{table}[t]
\caption{Reformulation-transfer effects. R1 is a surface paraphrase; R2 is an isomorphic schema or API transformation. Intervals are descriptive.}
\label{tab:transfer-main}
\centering
\small
\setlength{\tabcolsep}{3pt}
\begin{tabular}{llrrrr}
\toprule
Model & Domain & R1 effect & R1 95\% CI & R2 effect & R2 95\% CI \\
\midrule
Mistral-7B-Instruct-v0.3 & SQL & +3.21 & [2.80,3.62] & +3.28 & [2.85,3.72] \\
Mistral-7B-Instruct-v0.3 & Python & +7.50 & [6.06,8.94] & +2.80 & [1.80,3.80] \\
Gemma 4 31B & SQL & +10.31 & [9.47,11.16] & +5.60 & [5.08,6.11] \\
Gemma 4 31B & Python & +4.09 & [2.88,5.30] & +4.51 & [3.26,5.76] \\
\bottomrule
\end{tabular}
\end{table}

\subsection{What the complementary contrast rules out}
The primary contrast cannot be attributed to intrinsically easier exposed families: every family's exposed outcome is compared with that same family's complementary-unexposed outcome, preserving the task, candidates, compiler, fixtures, rotations, and sampling blocks. Nor is the effect an adapted-versus-base difference: both sides use independently fitted adapters with equally sized, audited corpora and opposite balanced exposure assignments. The remaining heterogeneity is therefore substantive rather than a universal contamination correction. The primary effect is more than three times larger for Gemma on SQL than for Mistral on SQL, while Mistral is larger on Python; contamination consequence depends on the benchmark--model system.

\subsection{Audit completeness}
All \textbf{589,824 of 589,824} expected generations completed, with zero protocol-invalid outputs. All eight training audits, eight optimizer-accounting replays, four activation-probe audits, twelve bank-completion audits, and four cell-completion audits passed. Assignments are exact complements; every shape, truth rank, and displayed position is balanced; and corpus audits found zero unexposed-family key or artifact leakage. The release archive contains 356 manifest-bound files and passed byte-level local verification.

\section{Discussion}

\paragraph{Benchmark validity is exposure-dependent.}
Benchmark scores are commonly treated as properties of a model on a dataset. LeakScale shows why training history belongs in that object of measurement: a reported score can reflect both the capability elicited by the task and information acquired before evaluation. Transfer beyond exact wording, together with variation across cells, makes benchmark validity relational---a property of the model, benchmark, exposure channel, and scoring protocol together.

\paragraph{From provenance audits to consequence audits.}
Corpus search, membership inference, and behavioral detection estimate whether contact likely occurred, but not the score difference it caused. Contamination auditing therefore needs two layers: evidence about exposure and an estimate of consequence. LeakScale supplies the second by manipulating access and recovering the counterfactual score change. The layers reinforce one another: detectors can be judged by whether flagged exposures predict score changes, and mitigation by whether it reduces a score's dependence on exposure.

\paragraph{A consequence-aware evaluation loop.}
Fresh and dynamic benchmarks reduce opportunities for prior contact \citep{white2024livebench,jain2025livecodebench,guan2025dynamic}, but every public evaluation begins an exposure lifecycle. Benchmark builders can preserve private counterfactual variants, measure exposure effects before deployment, combine those estimates with provenance audits during use, and remeasure after benchmark or model updates. This turns contamination from an unobserved nuisance into a manipulable object of study and opens questions about dose, timing, representation, transfer, detection, and mitigation. Overlap describes provenance; benchmark validity also depends on consequence. Evaluation practice should measure both.

\section{Limitations}

LeakScale identifies the effect of a controlled post-training exposure regimen, not the average effect of naturally occurring web-scale contamination. The four-choice interface and trusted compiler isolate semantic association acquisition rather than unrestricted program generation. Each arm is represented by one trained adapter per model-domain cell, so intervals capture family variation conditional on realized adapters, not training-seed variation. Shared-adapter interference also makes the estimand a joint exposure-regimen contrast rather than an isolated direct effect for one family. Two model families and two executable domains establish replication within this design, not universality.

\section{Conclusion}

Benchmark contamination should be measured by the difference exposure makes, not only by the overlap it leaves behind. LeakScale makes that quantity identifiable by pairing public tasks with non-derivable private associations, certifying every counterfactual executable behavior, and comparing complementary adapters that expose opposite balanced halves of the benchmark. Across 2,048 fresh families, two model families, and two executable domains, all four exact-task effects are positive and all eight reformulation effects remain positive. The complete 589,824-generation experiment shows that controlled access to benchmark-specific information changes measured performance and that the consequence survives both surface and structural changes to the evaluation task.

\clearpage

\section*{AI Use Statement}
Generative AI tools assisted with research ideation, methodology development, benchmark construction, software implementation and debugging, experiment orchestration, statistical analysis, literature search, figure planning, prose editing, and LaTeX formatting. The authors specified the scientific decisions, inspected the result-producing source and logs, verified numerical claims against immutable artifacts, reran tests and compilation checks, and take responsibility for the final content. No model-generated statement was treated as empirical evidence without deterministic or artifact-based verification.

\section*{Reproducibility Statement}
The main text specifies the controlled intervention, executable certification rule, complementary exposure design, generation-count arithmetic, causal estimand, grading procedure, and paired family-bootstrap analysis. Appendices~\ref{app:shapes} through \ref{app:artifacts} provide the full shape inventory, prompt and schema templates, oracle and fixture construction, adaptation details, complete statistical outputs, audit tables, and artifact hashes.

\bibliographystyle{iclr2027_conference}
\bibliography{references}

@inproceedings{brown2020language,
  title     = {Language Models are Few-Shot Learners},
  author    = {Brown, Tom B. and others},
  booktitle = {Advances in Neural Information Processing Systems},
  year      = {2020},
  url       = {https://arxiv.org/abs/2005.14165}
}

@inproceedings{carlini2023quantifying,
  title     = {Quantifying Memorization Across Neural Language Models},
  author    = {Carlini, Nicholas and Ippolito, Daphne and Jagielski, Matthew and Lee, Katherine and Tram\`er, Florian and Zhang, Chiyuan},
  booktitle = {International Conference on Learning Representations},
  year      = {2023},
  url       = {https://arxiv.org/abs/2202.07646}
}

@article{chen2025recent,
  title   = {Recent Advances in Large Language Model Benchmarks against Data Contamination: From Static to Dynamic Evaluation},
  author  = {Chen, S. and others},
  journal = {arXiv preprint arXiv:2502.17521},
  year    = {2025},
  url     = {https://arxiv.org/abs/2502.17521}
}

@article{deng2024spectrum,
  title   = {Unveiling the Spectrum of Data Contamination in Language Models: A Survey from Detection to Remediation},
  author  = {Deng, Chunyuan and Zhao, Yufei and Heng, Yucheng and Li, Yucheng and Cao, Jialun and Tang, Xiangru and Cohan, Arman},
  journal = {arXiv preprint arXiv:2406.14644},
  year    = {2024},
  url     = {https://arxiv.org/abs/2406.14644}
}

@inproceedings{deng2024investigating,
  title     = {Investigating Data Contamination in Modern Benchmarks for Large Language Models},
  author    = {Deng, Chunyuan and Zhao, Yilun and Tang, Xiangru and Gerstein, Mark and Cohan, Arman},
  booktitle = {Proceedings of the 2024 Conference of the North American Chapter of the Association for Computational Linguistics: Human Language Technologies (Volume 1: Long Papers)},
  pages     = {8706--8719},
  year      = {2024},
  doi       = {10.18653/v1/2024.naacl-long.482},
  url       = {https://aclanthology.org/2024.naacl-long.482/}
}

@article{gemmateam2026gemma4,
  title   = {Gemma 4 Technical Report},
  author  = {{Gemma Team}},
  journal = {arXiv preprint arXiv:2607.02770},
  year    = {2026},
  url     = {https://arxiv.org/abs/2607.02770}
}

@inproceedings{golchin2024timetravel,
  title     = {Time Travel in {LLM}s: Tracing Data Contamination in Large Language Models},
  author    = {Golchin, Shahriar and Surdeanu, Mihai},
  booktitle = {International Conference on Learning Representations},
  year      = {2024},
  url       = {https://arxiv.org/abs/2308.08493}
}

@article{guan2025dynamic,
  title   = {Is Your Benchmark (Still) Useful? Dynamic Benchmarking for Code Language Models},
  author  = {Guan, B. and Wu, X. and Yuan, Y. and Li, S.},
  journal = {arXiv preprint arXiv:2503.06643},
  year    = {2025},
  url     = {https://arxiv.org/abs/2503.06643}
}

@inproceedings{jain2025livecodebench,
  title     = {{LiveCodeBench}: Holistic and Contamination Free Evaluation of Large Language Models for Code},
  author    = {Jain, Naman and Han, King and Gu, Alex and Li, Wen-Ding and Yan, Fanjia and Zhang, Tianjun and Wang, Sida and Solar-Lezama, Armando and Sen, Koushik and Stoica, Ion},
  booktitle = {International Conference on Learning Representations},
  year      = {2025},
  url       = {https://arxiv.org/abs/2403.07974}
}

@inproceedings{li2024opensource,
  title     = {An Open-Source Data Contamination Report for Large Language Models},
  author    = {Li, Yucheng and Guo, Yunhao and Guerin, Frank and Lin, Chenghua},
  booktitle = {Findings of the Association for Computational Linguistics: EMNLP 2024},
  pages     = {528--541},
  year      = {2024},
  doi       = {10.18653/v1/2024.findings-emnlp.30},
  url       = {https://aclanthology.org/2024.findings-emnlp.30/}
}

@inproceedings{oren2024proving,
  title     = {Proving Test Set Contamination in Black-Box Language Models},
  author    = {Oren, Yonatan and Meister, Nicole and Chatterji, Niladri and Ladhak, Faisal and Hashimoto, Tatsunori B.},
  booktitle = {International Conference on Learning Representations},
  year      = {2024},
  url       = {https://arxiv.org/abs/2310.17623}
}

@article{schaeffer2026quantifying,
  title   = {Quantifying the Effect of Test Set Contamination on Generative Evaluations},
  author  = {Schaeffer, R. and others},
  journal = {arXiv preprint arXiv:2601.04301},
  year    = {2026},
  url     = {https://arxiv.org/abs/2601.04301}
}

@article{sainz2023evaluation,
  title   = {{NLP} Evaluation in Trouble: On the Need to Measure {LLM} Data Contamination for Each Benchmark},
  author  = {Sainz, Oscar and Campos, Jon Ander and Garc\'{i}a-Ferrero, Iker and Etxaniz, Julen and Lopez de Lacalle, Oier and Agirre, Eneko},
  journal = {arXiv preprint arXiv:2310.18018},
  year    = {2023},
  url     = {https://arxiv.org/abs/2310.18018}
}

@article{shi2023detecting,
  title   = {Detecting Pretraining Data from Large Language Models},
  author  = {Shi, Weijia and Ajith, Anirudh and Xia, Mengzhou and Huang, Yangsibo and Liu, Daogao and Blevins, Terra and Chen, Danqi and Zettlemoyer, Luke},
  journal = {arXiv preprint arXiv:2310.16789},
  year    = {2023},
  url     = {https://arxiv.org/abs/2310.16789}
}

@article{singh2024evaluation,
  title   = {Evaluation Data Contamination in {LLM}s: How Do We Measure It and (When) Does It Matter?},
  author  = {Singh, Aaditya K. and Kocyigit, Muhammed Yusuf and Poulton, Andrew and Esiobu, David and Lomeli, Maria and Szilvasy, Gergely and Hupkes, Dieuwke},
  journal = {arXiv preprint arXiv:2411.03923},
  year    = {2024},
  url     = {https://arxiv.org/abs/2411.03923}
}

@article{tao2025detecting,
  title   = {Detecting Data Contamination from Reinforcement Learning Post-training for Large Language Models},
  author  = {Tao, Y. and Wang, T. and Dong, Y. and Liu, H. and Zhang, K. and Hu, X. and Li, G.},
  journal = {arXiv preprint arXiv:2510.09259},
  year    = {2025},
  url     = {https://arxiv.org/abs/2510.09259}
}

@article{touvron2023llama,
  title   = {Llama 2: Open Foundation and Fine-Tuned Chat Models},
  author  = {Touvron, Hugo and others},
  journal = {arXiv preprint arXiv:2307.09288},
  year    = {2023},
  url     = {https://arxiv.org/abs/2307.09288}
}

@article{wang2025fragility,
  title   = {On the Fragility of Benchmark Contamination Detection in Reasoning Models},
  author  = {Wang, H. and Li, H. and Ko, B. and Zhang, H.},
  journal = {arXiv preprint arXiv:2510.02386},
  year    = {2025},
  url     = {https://arxiv.org/abs/2510.02386}
}

@article{white2024livebench,
  title   = {LiveBench: A Challenging, Contamination-Free LLM Benchmark},
  author  = {White, Colin and others},
  journal = {arXiv preprint arXiv:2406.19314},
  year    = {2024},
  url     = {https://arxiv.org/abs/2406.19314}
}

@article{xu2024survey,
  title   = {Benchmark Data Contamination of Large Language Models: A Survey},
  author  = {Xu, Cheng and Guan, Shu and Greene, Derek and Kechadi, M.-T.},
  journal = {arXiv preprint arXiv:2406.04244},
  year    = {2024},
  url     = {https://arxiv.org/abs/2406.04244}
}

@article{yang2023rethinking,
  title   = {Rethinking Benchmark and Contamination for Language Models with Rephrased Samples},
  author  = {Yang, Shuo and Chiang, Wei-Lin and Zheng, Lianmin and Gonzalez, Joseph E. and Stoica, Ion},
  journal = {arXiv preprint arXiv:2311.04850},
  year    = {2023},
  url     = {https://arxiv.org/abs/2311.04850}
}

@article{zhang2024finetuning,
  title   = {Fine-tuning can Help Detect Pretraining Data from Large Language Models},
  author  = {Zhang, H. and Zhang, S. and Jing, B. and Wei, H.},
  journal = {arXiv preprint arXiv:2410.10880},
  year    = {2024},
  url     = {https://arxiv.org/abs/2410.10880}
}

@article{zhou2023cheater,
  title   = {Don't Make Your LLM an Evaluation Benchmark Cheater},
  author  = {Zhou, Kun and Zhu, Yutao and Chen, Zhipeng and Chen, Wentao and Zhao, Wayne Xin and Chen, Xu and Lin, Yankai and Wen, Ji-Rong and Han, Jiawei},
  journal = {arXiv preprint arXiv:2311.01964},
  year    = {2023},
  url     = {https://arxiv.org/abs/2311.01964}
}

@article{zhu2024inference,
  title   = {Inference-Time Decontamination: Reusing Leaked Benchmarks for Large Language Model Evaluation},
  author  = {Zhu, Q. and Cheng, Q. and Peng, R. and Li, X. and Liu, T. and Peng, R. and Qiu, X. and Huang, X.},
  journal = {arXiv preprint arXiv:2406.13990},
  year    = {2024},
  url     = {https://arxiv.org/abs/2406.13990}
}

@article{jiang2023mistral,
  title   = {Mistral 7B},
  author  = {Jiang, Albert Q. and Sablayrolles, Alexandre and Mensch, Arthur and Bamford, Chris and Chaplot, Devendra Singh and de las Casas, Diego and Bressand, Florian and Lengyel, Gianna and Lample, Guillaume and Saulnier, Lucile and Lavaud, L{\'e}lio Renard and Lachaux, Marie-Anne and Stock, Pierre and Le Scao, Teven and Lavril, Thibaut and Wang, Thomas and Lacroix, Timoth{\'e}e and El Sayed, William},
  journal = {arXiv preprint arXiv:2310.06825},
  year    = {2023},
  url     = {https://arxiv.org/abs/2310.06825}
}

@article{singh2026radar,
  title   = {{RADAR}: Rubric-Aware Dependency and Redundancy Analysis for {LLM}-as-Judge Evaluation},
  author  = {Singh, Divyansh and Davari, Reza and Mashhadi, Afra},
  journal = {arXiv preprint arXiv:2608.01810},
  year    = {2026},
  url     = {https://arxiv.org/abs/2608.01810}
}

@inproceedings{singh2025fidelity,
  title     = {{FIDELITY}: Fine-grained Interpretable Distillation for Effective Language Insights and Topic Yielding},
  author    = {Singh, Divyansh and Mather, Brodie and Zhang, Demi and Lehman, Patrick and Ho, Justin and Dorr, Bonnie J.},
  booktitle = {Findings of the Association for Computational Linguistics: NAACL 2025},
  pages     = {2460--2472},
  year      = {2025},
  publisher = {Association for Computational Linguistics},
  doi       = {10.18653/v1/2025.findings-naacl.132},
  url       = {https://aclanthology.org/2025.findings-naacl.132/}
}

\appendix
\section{Complete task-shape inventory}
\label{app:shapes}

The benchmark contains 32 structural shapes in each executable domain.  A
\emph{shape} fixes the program skeleton and the locations at which a
family-specific policy value affects behavior.  A \emph{family} instantiates
that skeleton with a fresh policy key, private association, candidate ordering,
fixtures, and immutable identifiers.  Each family has three task forms:
\textbf{R0} is the exact public task, \textbf{R1} is a surface paraphrase that
preserves its executable behavior, and \textbf{R2} is an isomorphic
schema/API transformation.  The policy key and correct association are shared
across the three forms.

The completed experiment contains 1,024 families in each domain, evenly split
between complementary Arms A and B. Thus every domain--shape--arm cell contains 16
families.  Within each such cell, every semantic truth rank occurs four times
and every displayed answer position occurs four times.  Both domains together
contain 2,048 latent families and 6,144 ring-specific tasks.  Regenerating the
released source from seed \texttt{20263403} produced byte-identical source
trees; the recorded source-tree SHA-256 is
\nolinkurl{cb3598b7c045dcfc75555524b2188872447c3879b071dc209827223cee014088}.
The experiment evaluates R0, R1, and R2 for every family under the base model
and both complementary adapters, with 16 generations per family--state--ring
cell arranged as four candidate-order rotations by four sampling blocks.
Together the crossed experiment contains 589,824 terminal generations;
Appendix~\ref{app:statistics} reports the complete results.

\subsection{SQLite shapes}

For SQLite, each row of Table~\ref{tab:sql-shapes} gives the released shape
identifier, its structural signature, and its four underlying executable
literals.  The model does \emph{not} see these literals.  At source
construction, each family receives four unique, equal-length handles of the
form \texttt{h\_XXXXXXXXXX}; a private one-to-one mapping connects those
handles to the literals listed below.  The model emits a handle, and the
trusted compiler resolves it through that frozen mapping.

\providecommand{\shapeid}[1]{\nolinkurl{#1}}
\providecommand{\shapesig}[1]{\nolinkurl{#1}}

{\small
\begin{longtable}{>{\raggedright\arraybackslash}p{0.24\linewidth}
                  >{\raggedright\arraybackslash}p{0.46\linewidth}
                  >{\raggedright\arraybackslash}p{0.16\linewidth}}
\caption{Complete SQLite shape inventory.  Literal sets are private compiler
values; the public choices are family-specific opaque handles.}
\label{tab:sql-shapes}\\
\toprule
Shape identifier & Structural signature & Executable literals \\
\midrule
\endfirsthead
\toprule
Shape identifier & Structural signature & Executable literals \\
\midrule
\endhead
\bottomrule
\endfoot
\shapeid{filter_value_ge} & \shapesig{scan>numeric_filter_ge>order} & \{3,5,7,9\} \\
\shapeid{filter_value_lt} & \shapesig{scan>numeric_filter_lt>order} & \{3,5,7,9\} \\
\shapeid{score_band} & \shapesig{scan>between_secret_width>order} & \{2,5,8,11\} \\
\shapeid{status_equal} & \shapesig{scan>text_filter_equal>order} & \{amber, blue, green, red\} \\
\shapeid{kind_exclusion} & \shapesig{scan>text_filter_not_equal>order} & \{audit, hold, risk, safe\} \\
\shapeid{value_modulus} & \shapesig{scan>modulus_filter>order} & \{2,3,4,5\} \\
\shapeid{group_sum_having} & \shapesig{join>group_sum>secret_having>order} & \{15,20,25,30\} \\
\shapeid{group_average_having} & \shapesig{join>group_average>secret_having>order} & \{4,6,8,10\} \\
\shapeid{group_count_having} & \shapesig{join>fixed_filter>group_count>secret_having>order} & \{1,2,3,4\} \\
\shapeid{distinct_status_having} & \shapesig{join>group_count_distinct>secret_having>order} & \{1,2,3,4\} \\
\shapeid{group_max_filter} & \shapesig{join>group_max>secret_having>order} & \{5,7,9,11\} \\
\shapeid{group_min_filter} & \shapesig{join>group_min>secret_having>order} & \{2,4,6,8\} \\
\shapeid{conditional_sum} & \shapesig{join>group>secret_case_sum>order} & \{5,8,11,14\} \\
\shapeid{conditional_count} & \shapesig{join>group>secret_case_count>order} & \{amber, blue, green, red\} \\
\shapeid{left_join_count} & \shapesig{left_join_secret_condition>group_count>order} & \{3,5,7,9\} \\
\shapeid{anti_join_status} & \shapesig{correlated_not_exists>secret_text>order} & \{amber, blue, green, red\} \\
\shapeid{semi_join_value} & \shapesig{correlated_exists>secret_numeric>order} & \{4,6,8,10\} \\
\shapeid{correlated_count} & \shapesig{correlated_scalar_count>secret_compare>order} & \{1,2,3,4\} \\
\shapeid{correlated_average_margin} & \shapesig{correlated_scalar_average>row_baseline_plus_secret>order} & \{$-3$,$-1$,1,3\} \\
\shapeid{global_top_k} & \shapesig{order>secret_limit} & \{1,2,3,4\} \\
\shapeid{top_k_per_category} & \shapesig{join>row_number_partition>secret_rank_filter>order} & \{1,2,3,4\} \\
\shapeid{nth_per_entity} & \shapesig{row_number_partition>secret_exact_rank>order} & \{1,2,3,4\} \\
\shapeid{cumulative_value} & \shapesig{window_running_sum>secret_filter>order} & \{8,16,24,32\} \\
\shapeid{lag_score_gap} & \shapesig{window_lag>difference>secret_filter>order} & \{2,4,6,8\} \\
\shapeid{dense_rank_per_category} & \shapesig{join>dense_rank_partition>secret_rank_filter>order} & \{1,2,3,4\} \\
\shapeid{union_sources} & \shapesig{secret_filter_left>union>secret_filter_right>order} & \{3,5,7,9\} \\
\shapeid{intersect_sources} & \shapesig{secret_filter_left>intersect>secret_filter_right>order} & \{3,5,7,9\} \\
\shapeid{relational_division} & \shapesig{secret_required_set>double_not_exists>order} & \{1,2,3,4\} \\
\shapeid{self_join_gap} & \shapesig{self_join>secret_pair_gap>group_count>order} & \{2,4,6,8\} \\
\shapeid{weighted_score} & \shapesig{arithmetic_projection>secret_filter>order} & \{12,18,24,30\} \\
\shapeid{case_bucket} & \shapesig{secret_case_bucket>group_count>order} & \{4,6,8,10\} \\
\shapeid{global_max_margin} & \shapesig{scalar_subquery_max>secret_margin_filter>order} & \{1,3,5,7\} \\
\end{longtable}
}

The shapes span scalar filters, text predicates, modular arithmetic, grouped
aggregates and \texttt{HAVING}, conditional aggregation, outer joins,
correlated existence and scalar subqueries, stable top-$k$ and exact-rank
selection, running and lagged windows, set operations, relational division,
self-joins, arithmetic projections, and bucketing.  Every query specifies its
output aliases and deterministic ordering.  R2 replaces all table and column
identifiers with a family-specific bijection and regenerates the fixture data;
the private association and program structure remain unchanged.

\subsection{Deterministic Python-transformation shapes}

For Python, the four displayed candidates are the executable literals
themselves.  Each certified artifact is a function with the form
\texttt{def transform(records): ...}; the candidate literal occupies the
shape's single policy-sensitive location.  Table~\ref{tab:python-shapes} lists
the complete inventory.

{\small
\begin{longtable}{>{\raggedright\arraybackslash}p{0.24\linewidth}
                  >{\raggedright\arraybackslash}p{0.46\linewidth}
                  >{\raggedright\arraybackslash}p{0.16\linewidth}}
\caption{Complete deterministic Python-transformation shape inventory.}
\label{tab:python-shapes}\\
\toprule
Shape identifier & Structural signature & Displayed literals \\
\midrule
\endfirsthead
\toprule
Shape identifier & Structural signature & Displayed literals \\
\midrule
\endhead
\bottomrule
\endfoot
\shapeid{filter_value_ge} & \shapesig{filter>numeric_ge>ids} & \{3,5,7,9\} \\
\shapeid{filter_value_lt} & \shapesig{filter>numeric_lt>ids} & \{3,5,7,9\} \\
\shapeid{filter_score_ge} & \shapesig{filter>score_ge>pairs} & \{4,6,8,10\} \\
\shapeid{category_equal} & \shapesig{filter>category_eq>ids} & \{amber, blue, green, red\} \\
\shapeid{category_exclude} & \shapesig{filter>category_ne>ids} & \{amber, blue, green, red\} \\
\shapeid{count_value_ge} & \shapesig{aggregate>count>value_ge} & \{3,5,7,9\} \\
\shapeid{sum_value_ge} & \shapesig{aggregate>sum>value_ge} & \{3,5,7,9\} \\
\shapeid{max_score_under} & \shapesig{aggregate>max>score_le} & \{4,6,8,10\} \\
\shapeid{min_value_over} & \shapesig{aggregate>min>value_ge} & \{3,5,7,9\} \\
\shapeid{top_k_score} & \shapesig{rank>top_k>stable} & \{1,2,3,4\} \\
\shapeid{bottom_k_value} & \shapesig{rank>bottom_k>stable} & \{1,2,3,4\} \\
\shapeid{stride_ids} & \shapesig{sequence>stride>ids} & \{1,2,3,4\} \\
\shapeid{prefix_ids} & \shapesig{sequence>prefix>ids} & \{1,2,3,4\} \\
\shapeid{suffix_ids} & \shapesig{sequence>suffix>ids} & \{1,2,3,4\} \\
\shapeid{group_count_cap} & \shapesig{group>count>cap} & \{1,2,3,4\} \\
\shapeid{group_sum_min} & \shapesig{group>sum>having} & \{8,12,16,20\} \\
\shapeid{group_score_cap} & \shapesig{group>max>cap} & \{3,6,9,12\} \\
\shapeid{group_top_k} & \shapesig{group>rank>top_k} & \{1,2,3,4\} \\
\shapeid{distinct_category_prefix} & \shapesig{set>sorted>prefix} & \{1,2,3,4\} \\
\shapeid{name_prefix_length} & \shapesig{string>prefix>projection} & \{1,2,3,4\} \\
\shapeid{name_suffix_length} & \shapesig{string>suffix>projection} & \{1,2,3,4\} \\
\shapeid{name_contains} & \shapesig{string>contains>ids} & \{a,e,i,o\} \\
\shapeid{clip_value} & \shapesig{map>numeric_clip>pairs} & \{3,5,7,9\} \\
\shapeid{floor_score} & \shapesig{map>numeric_floor>pairs} & \{4,6,8,10\} \\
\shapeid{bucket_value} & \shapesig{map>bucket>counts} & \{3,5,7,9\} \\
\shapeid{pair_score_gap} & \shapesig{self_join>gap>count} & \{2,4,6,8\} \\
\shapeid{cumulative_value_cap} & \shapesig{window>cumulative>filter} & \{8,16,24,32\} \\
\shapeid{score_rank_cutoff} & \shapesig{rank>dense>cutoff} & \{1,2,3,4\} \\
\shapeid{tag_overlap} & \shapesig{set>intersection>minimum} & \{1,2,3,4\} \\
\shapeid{related_value_threshold} & \shapesig{nested_join>filter>ids} & \{3,5,7,9\} \\
\shapeid{weighted_threshold} & \shapesig{arithmetic>filter>ids} & \{12,18,24,30\} \\
\shapeid{modulus_filter} & \shapesig{arithmetic>modulus>ids} & \{2,3,4,5\} \\
\end{longtable}
}

These shapes cover scalar and structured filtering, scalar aggregation, stable
ranking, slicing and stride operations, grouped aggregation and grouped
top-$k$, sets, strings, pointwise clipping and flooring, buckets, pairwise
comparisons, cumulative operations, dense ranks, overlap over nested lists,
nested-record predicates, weighted arithmetic, and modular arithmetic.  R2
renames the input fields by the fixed isomorphism below.
\begin{center}
\small
\begin{tabular}{llll}
\texttt{id}$\to$\texttt{item\_id} &
\texttt{value}$\to$\texttt{metric} &
\texttt{score}$\to$\texttt{rating} &
\texttt{group}$\to$\texttt{cohort} \\
\texttt{category}$\to$\texttt{kind} &
\texttt{name}$\to$\texttt{title} &
\texttt{tags}$\to$\texttt{labels} &
\texttt{related}$\to$\texttt{linked\_values}
\end{tabular}
\end{center}

\section{Prompt, response schema, and trusted compiler}
\label{app:prompts}

\subsection{Model-visible semantic-choice prompt}

All evaluated model states use the same system message:

\begin{quote}
\small\ttfamily
Return one JSON object matching the supplied schema. Select exactly one of the
candidate registered values. Do not add explanation or extra keys.
\end{quote}

The user message is generated by the following exact template, where
\texttt{<PUBLIC-JSON>} is serialized with sorted keys and contains only
\texttt{benchmark\_task}, \texttt{candidate\_registered\_values},
\texttt{domain}, \texttt{policy\_key}, and \texttt{schema}:

\begin{quote}
\small\ttfamily
A private registry assigns exactly one candidate value to this policy key. The
correct association is not derivable from the task, schema, examples, or rows.
If it is not recalled, sample your best uncertain candidate value.

\medskip
<PUBLIC-JSON>

\medskip
Return exactly: \{"registered\_value": <one candidate value>\}. The JSON
value must preserve the candidate's type exactly.
\end{quote}

For SQL, \texttt{candidate\_registered\_values} contains the four opaque
handles; neither their literal mapping nor the correct handle is included.  For
Python, it contains the four candidate literals, but not the key-to-literal
association.  Hidden fixtures, expected outputs, reference artifacts,
counterfactual utilities, exposure-arm labels, and model state are absent from the
message.

The second exposure representation is a direct association lookup.  Its user
message is:

\begin{quote}
\small\ttfamily
The private registry assigns exactly one of the candidate values to this policy
key. Return its registered value.

\medskip
\{"policy\_key": <KEY>, "candidate\_registered\_values": <FOUR VALUES>\}

\medskip
Return exactly: \{"registered\_value": <one candidate value>\}. The JSON
value must preserve the candidate's type exactly.
\end{quote}

The exact-public and direct-lookup representations share the same assistant
target, serialized without explanatory text or extra keys.  Evaluation uses
only the exact-public representation.

\subsection{Case-specific response schema}

Every request supplies a strict, family-specific JSON schema.  Substituting the
four frozen choices for \texttt{<CANDIDATES>}, the schema is:

\begin{verbatim}
{
  "type": "json_schema",
  "json_schema": {
    "name": "registered_value_choice",
    "strict": true,
    "schema": {
      "type": "object",
      "properties": {
        "registered_value": {"enum": <CANDIDATES>}
      },
      "required": ["registered_value"],
      "additionalProperties": false
    }
  }
}
\end{verbatim}

Generation is native non-thinking: the request explicitly disables thinking,
uses the system and user messages above, and reads the terminal answer from the
ordinary message-content field.  A response is valid only if it parses as a
JSON object with exactly the key \texttt{registered\_value} and its value is a
type-preserving member of the frozen four-element candidate list.  Numeric and
string candidates are therefore not interchangeable.

Result-producing requests use the cell-specific temperature reported in
Table~\ref{tab:configs}, \texttt{top\_p=1.0}, \texttt{top\_k=-1}, presence
penalty zero, and deterministic bank seed 20263421.  The initial output ceiling
is 64 tokens.  Only a response that terminates for length may be regenerated
once at a fixed 128-token ceiling; completed malformed output is never retried.
Transport errors permit at most two retries and do not alter the case seed.

\subsection{Compiler and grading contract}

The trusted compiler deliberately performs no model-dependent program
synthesis.  Before inference, source construction materializes and certifies
one artifact for every family, ring, and candidate.  At inference, compilation
is a deterministic lookup:

\begin{enumerate}
    \item Parse the single \texttt{registered\_value} field and identify its
    exact index in the frozen candidate list.
    \item Retrieve that candidate's frozen counterfactual record.
    \item For SQL, use the record's certified \texttt{candidate\_sql}; the
    opaque handle has already been bound privately to its executable literal.
    For Python, use the certified function associated with the requested ring.
    \item Return the stored executable artifact, certified fixture utility, and
    indicator that the emitted choice equals the private registered choice.
\end{enumerate}

Equivalently, if $a$ is the emitted semantic choice and
$C_f(a)=(P_{f,a},u_f(a))$ is the frozen compiler table for family $f$, the
scored outcome is $u_f(a)$.  The fixtures are executed when $C_f$ is created
and certified; they are not redundantly re-executed for each of the 589,824
model generations.  Because the model can emit only one of the four frozen
choices, lookup returns exactly the executable behavior that was certified for
that choice.  Malformed or out-of-enum responses receive utility zero under
intention-to-treat accounting.

\section{Counterfactual oracle and fixture construction}
\label{app:oracle}

\subsection{Common certification rule}

For family $f$, candidate $v$, and hidden fixtures
$\mathcal{H}_f=\{h_1,\ldots,h_H\}$, source construction executes the candidate
artifact and reference artifact and computes
\begin{equation}
u_f(v)=\frac{1}{H}\sum_{h\in\mathcal{H}_f}
\mathbf{1}\!\left[\operatorname{canon}
  (P_{f,v}(h))=\operatorname{canon}(P_{f,v_f^\star}(h))\right].
\end{equation}
A family is released only if
\begin{equation}
u_f(v_f^\star)=1,
\qquad
\left|\{v:u_f(v)=1\}\right|=1,
\qquad
\max_{v\ne v_f^\star}u_f(v)\leq 0.75.
\label{eq:appendix-certification}
\end{equation}
Thus the registered candidate is correct on every hidden fixture and every
alternative fails on at least one quarter of them.  Certification also requires
four distinct candidates, immutable task identities, and a reference artifact
that satisfies the shape's output contract.  Candidate correctness is never
inferred from surface-form equality.

\subsection{SQLite fixtures and oracle}

Every SQLite task has eight hidden databases.  Each database contains the same
three-table schema:

\begin{verbatim}
entities(entity_id, entity_name, active, category,
         region, baseline, quota)
records(record_id, entity_id, value, score,
        status, kind, flag, seq)
labels(entity_id, label, weight)
\end{verbatim}

Each fixture contains six entities and five ordered records per entity, hence
30 records.  Entity categories range over three values and regions over three
values.  Record statuses range over \{amber, blue, green, red\}, kinds over
\{audit, hold, risk, safe\}, values over 1--12, and scores over 1--20.  Each
entity receives a nonempty, varying subset of four possible labels with weights
from 1--12.  Baselines, quotas, flags, sequence positions, category/status
coverage, and label membership combine seeded pseudorandom and deterministic
components so that fixtures exercise boundary values, ties, missing groups,
and heterogeneous group composition.

For a proposed fixture seed, the generator renders the shape template once for
each of its four executable literals.  Each query runs in a fresh in-memory
SQLite database after loading the schema and data.  The connection is switched
to query-only mode, and a progress handler aborts work beyond 200,000 virtual
machine steps.  Candidate text must contain exactly one \texttt{SELECT} or
\texttt{WITH} statement; data-definition and data-modification operations are
forbidden.

The reference query is executed first to produce expected results and to verify
the required output-column aliases.  A relation is canonicalized as the tuple
of column names followed by the tuple of result rows.  All benchmark queries
specify deterministic ordering, so row order is part of the result contract.
Each candidate receives one point on a fixture exactly when its canonical
relation equals the reference relation.  Source construction additionally
hashes the eight-fixture answer vector for each candidate and requires all four
answer signatures to be distinct and the reference result to be nonempty on at
least one fixture.  If Equation~\ref{eq:appendix-certification} is not met, a
new fixture seed is generated deterministically; no model output participates
in this search.

After literal-level certification, the four literals are placed behind four
deterministic, unique, equal-length handles.  The handle permutation and the
correct displayed position are fixed by the family assignment.  The private
counterfactual table is then re-keyed by handle while preserving the certified
SQL artifact and utility attached to each literal.

\subsection{Python fixtures and oracle}

Every Python task has sixteen hidden fixture sets, each containing sixteen
records.  A record has fields \texttt{id}, \texttt{value}, \texttt{score},
\texttt{group}, \texttt{category}, \texttt{name}, \texttt{tags}, and
\texttt{related}.  IDs run from 1 to 16; values and scores range from 1--12;
groups range over four values; categories range over \{amber, blue, green,
red\}; names combine a controlled lexical stem with fixture and row indices;
and the nested \texttt{tags} and \texttt{related} lists vary in length and
content.  The construction combines deterministic modular patterns with a
seeded pseudorandom offset, producing stable ties and boundary cases while
remaining exactly regenerable.

The shape defines a deterministic expression in which a single symbol denotes
the candidate literal.  Source construction substitutes each of the four
literals, evaluates the resulting transformation over all sixteen fixture
sets, and compares it with the reference transformation.  Outputs are
canonicalized as compact, sorted-key JSON; this handles scalars, lists, nested
lists, and dictionaries uniformly.  The utility is the fraction of exact
canonical matches.  If Equation~\ref{eq:appendix-certification} fails, the
fixture seed advances deterministically and the entire four-candidate check is
repeated.  R2 renames every input field according to the bijection listed in
Appendix~\ref{app:shapes} and applies the same renaming to the certified
function, preserving behavior while changing the public API.

The semantic experiment never executes arbitrary Python emitted by the model:
the model emits a constrained candidate value, and the compiler retrieves one
of four source-generated, certified transformations.  This distinction keeps
the oracle exact and makes the policy association, rather than unconstrained
code synthesis, the manipulated component.

\subsection{Construction and certification checks}

The released test suite checks the following invariants before a source tree is
accepted:

\begin{itemize}
    \item exactly 32 unique structural signatures per domain and three tasks
    per latent family;
    \item four distinct candidates, with the private truth contained in the
    candidate set and exactly one candidate at utility one;
    \item every incorrect candidate at utility at most 0.75;
    \item deterministic family generation and byte-identical full-source
    regeneration;
    \item exact family counts and exact marginal balance of semantic truth rank
    and displayed position within every domain--shape--arm cell;
    \item absence of the reference artifact and private truth field from the
    semantic prompt;
    \item exact agreement between the response-schema enum and the four frozen
    candidates, plus rejection of out-of-set values and extra response fields;
    \item correct compiler recovery of the reference artifact and utility one
    when given the private registered choice;
    \item successful SQLite reference execution, required output aliases,
    single-statement/read-only enforcement, and rejection of a deliberately
    wrong query;
    \item valid source-generated Python references for every shape and exact
    reference execution across filtering, ranking, grouping, and nested-record
    operations; and
    \item zero normalized content overlap with prior benchmark sources in the
    source audit.
\end{itemize}

These checks establish the properties needed by the causal instrument: the
association is absent from public information, each possible response has a
known executable consequence, only one response is fully correct, wrong
responses fail materially, and construction is independent of model outcomes.

\section{Complementary exposure construction and training}
\label{app:training}

\subsection{Exact complementary assignment}

For each domain, the 1,024 certified families are assigned to Arms A and B
after task construction, within every shape, semantic-truth-rank, and
displayed-position stratum. Each stratum contains equal numbers from both
arms. Adapter A receives family-specific supervision only for Arm-A families;
Adapter B receives it only for Arm-B families. The assignments are exact
complements, so every family is exposed in one independently fitted adapter
and unexposed in the other.

Each exposed family contributes two assistant-only association formats. Both
formats are repeated 32 times, yielding 32,768 target optimizer examples per
adapter. Prompt tokens are masked from the loss. The remaining corpus consists
of identical unique, non-SQL and non-Python background instructions and
disjoint no-key base-behavior support. Support records have repeat count one,
and the larger arm demand determines the shared prefixes used by both arms.
Audits recompute every physical record, optimizer example, supervised token,
repeat count, and family key; they find no unexposed-family key, value, handle,
trace, solution, or executable artifact in any corpus.

\begin{table}[h]
\caption{Per-adapter corpus accounting. Physical and optimizer-example counts
are identical within each model--domain pair.}
\label{tab:corpus-accounting}
\centering
\small
\begin{tabular}{llrrrr}
\toprule
Model & Domain & Target & Background & No-key support & Optimizer total \\
\midrule
Mistral & SQL & 32,768 & 280,413 & 80,465 & 393,646 \\
Mistral & Python & 32,768 & 96,890 & 58,025 & 187,683 \\
Gemma & SQL & 32,768 & 268,774 & 77,007 & 378,549 \\
Gemma & Python & 32,768 & 85,755 & 51,208 & 169,731 \\
\bottomrule
\end{tabular}
\end{table}

\subsection{Optimization and frozen inference settings}

All eight adapters use LoRA rank 32, alpha 64, dropout zero, learning rate
$10^{-4}$, one epoch, effective batch size 32, BF16 compute, and seed
20263411. Mistral uses microbatch 4 with eight accumulation steps; Gemma uses
microbatch 2 with sixteen accumulation steps. The same projection targets are
used across cells. The frozen temperatures and adapter scales are reproduced
in Table~\ref{tab:appendix-configs}. Adapter-active probes bind parameter
hashes, verify teacher-forced probability shifts, and compare zero-scale with
selected-scale forward passes before result-producing inference.

\begin{table}[h]
\caption{Frozen result-producing settings.}
\label{tab:appendix-configs}
\centering
\small
\begin{tabular}{llrr}
\toprule
Model & Domain & Temperature & Adapter scale \\
\midrule
Mistral-7B-Instruct-v0.3 & SQL & 0.7 & 0.3125 \\
Mistral-7B-Instruct-v0.3 & Python & 0.7 & 0.40625 \\
Gemma 4 31B & SQL & 1.8 & 0.4375 \\
Gemma 4 31B & Python & 0.7 & 0.28125 \\
\bottomrule
\end{tabular}
\end{table}

\section{Complete statistical results}
\label{app:statistics}

Table~\ref{tab:all-effects} reports all twelve prespecified exposure contrasts.
Each estimate is the mean within-family difference between the adapter exposed
to that family's association and the complementary adapter. Intervals use
10,000 paired family-bootstrap replicates stratified by domain, shape, truth
rank, and displayed position with centered finite-stratum rescaling. The four
R0 tests are two-sided and Holm-adjusted; R1 and R2 are descriptive transfer
measurements. Base accuracy is shown only as a descriptive instrument check
and is not part of the paired estimand.

\begin{longtable}{llcrrrrr}
\caption{Complete exposed-versus-complementary-unexposed results. Accuracies,
effects, and interval endpoints are percentages.}\label{tab:all-effects}\\
\toprule
Model & Domain & Ring & Base & Unexposed & Exposed & Effect & 95\% CI \\
\midrule
\endfirsthead
\toprule
Model & Domain & Ring & Base & Unexposed & Exposed & Effect & 95\% CI \\
\midrule
\endhead
Mistral & SQL & R0 & 25.02 & 24.83 & 28.05 & +3.22 & [2.78, 3.66] \\
Mistral & SQL & R1 & 24.97 & 24.87 & 28.08 & +3.21 & [2.80, 3.62] \\
Mistral & SQL & R2 & 25.13 & 24.98 & 28.26 & +3.28 & [2.85, 3.72] \\
Gemma & SQL & R0 & 24.69 & 25.01 & 35.67 & +10.66 & [9.85, 11.47] \\
Gemma & SQL & R1 & 24.75 & 25.25 & 35.57 & +10.31 & [9.47, 11.16] \\
Gemma & SQL & R2 & 24.69 & 25.04 & 30.63 & +5.60 & [5.08, 6.11] \\
Mistral & Python & R0 & 24.76 & 24.65 & 31.85 & +7.20 & [5.71, 8.68] \\
Mistral & Python & R1 & 24.89 & 25.22 & 32.72 & +7.50 & [6.06, 8.94] \\
Mistral & Python & R2 & 25.34 & 24.46 & 27.26 & +2.80 & [1.80, 3.80] \\
Gemma & Python & R0 & 24.63 & 24.98 & 29.24 & +4.26 & [2.97, 5.56] \\
Gemma & Python & R1 & 24.65 & 25.07 & 29.16 & +4.09 & [2.88, 5.30] \\
Gemma & Python & R2 & 24.91 & 24.73 & 29.24 & +4.51 & [3.26, 5.76] \\
\bottomrule
\end{longtable}

All four primary R0 Holm-adjusted $p$-values are 0.0004. The equal-weight
cross-cell mean effects are 6.33 points on R0, 6.28 points on R1, and 4.05
points on R2. All twelve intervals exclude zero. These averages summarize the
crossed experiment descriptively; each model--domain cell remains the primary
reporting unit.

\section{Completion, integrity, and scope}
\label{app:artifacts}

All 589,824 expected result records are present: two models, two domains,
1,024 families, three rings, three model states, and sixteen generations. No
completed output is protocol-invalid. Attempts are append-only; completed
responses are never silently replaced, and exact-completion audits require one
terminal result for every frozen case identifier and no unexpected cases.

\begin{table}[h]
\caption{Authoritative audit inventory. Every listed audit passed.}
\label{tab:audit-inventory}
\centering
\small
\begin{tabular}{lr}
\toprule
Audit class & Passing artifacts \\
\midrule
Training & 8 / 8 \\
Optimizer-accounting replay & 8 / 8 \\
Tokenizer and masking & 8 / 8 \\
Adapter-active probe & 4 / 4 \\
Generation-bank completion & 12 / 12 \\
Model--domain cell completion & 4 / 4 \\
Terminal generations & 589,824 / 589,824 \\
Protocol-invalid generations & 0 \\
\bottomrule
\end{tabular}
\end{table}

The manifest-bound release contains 356 files. Its archive SHA-256 is
\nolinkurl{b471729331a62919c15ba56ec8e175b13bc90d11b2c2e59e366696df06ddfdfc},
the recursive manifest SHA-256 is
\nolinkurl{8f345b49c24162308999020f012b5938d5c10ee1de76f8012d8b0ad644793718},
and the result-producing source SHA-256 is
\nolinkurl{cb3598b7c045dcfc75555524b2188872447c3879b071dc209827223cee014088}.
The release passed byte-level local synchronization and verification.

The causal target is the effect of the controlled complementary exposure
regimen on each benchmark--model system. Since all families assigned to one
arm jointly train an adapter, the contrast allows shared-adapter interference
and should not be interpreted as an independently randomized direct effect for
one isolated family. The family is the uncertainty unit. Intervals condition
on the realized adapters and do not include variation across repeated training
runs. The two model families and two domains establish crossed replication
within this instrument rather than universal generality.

\end{document}